\documentclass{IEEEtran}

\usepackage[T1]{fontenc}
\usepackage[utf8]{inputenc}
\usepackage[english]{babel}

\usepackage{amsmath}
\usepackage{amssymb}
\usepackage{amsthm}

\usepackage{graphicx}
\usepackage{booktabs}
\usepackage{array}
\usepackage{multirow}

\usepackage{algorithm}
\usepackage{algpseudocode}

\usepackage[hidelinks]{hyperref}

\usepackage{url}

\title{
Conceptual full-vehicle modeling of a modular all-wheel-drive robotic ADS test platform as a foundation for a digital twin architecture
}

\author{
Felix Florian Schwald$^{1}$,
Martin Meiners$^{1,2,*}$,
Michael Viehof$^{1}$,
Erik Hultman$^{2}$,
Edwin Kamau$^{1}$
\\[4pt]
$^{1}$TH Köln, Germany
\\
$^{2}$Uppsala University, Sweden
\\[2pt]
$^{*}$Corresponding author:
\href{mailto:martin.meiners@th-koeln.de}
{martin.meiners@th-koeln.de}
}

\begin{document}

\maketitle

\begin{abstract}


This paper presents the development of an initial full-vehicle simulation model of the PixKit modular electric test platform in IPG CarMaker. The objective is to create a physics-based and extendable model structure that can serve as a low-level foundation for future digital twin applications. The reference vehicle is a skateboard chassis combining unconventional vehicle characteristics such as four independent wheel-hub motors and rear axle steering capable of high steering angles. Since complete measurement data of the vehicle is not yet available, the model parameterization is based on a combination of manufacturer documentation, derived values and engineering assumptions. The modeling approach focuses on the system-level integration of the main vehicle subsystems rather than on a fully validated representation.
The resulting model was assessed through basic driving scenarios with different steering inputs and configurations. The simulations showed stable model behavior and physically plausible interaction between the subsystems. The all-wheel-drive configuration produced symmetric wheel responses in longitudinal driving, while counter-phase rear axle steering increased the yaw response and reduced the simulated turning circle diameter compared with front axle steering only. The circular driving scenario further indicated bounded roll and slip behavior. Overall, the implemented CarMaker model provides a traceable and functional simulation basis for the PixKit vehicle. It can support future model refinement, calibration with real vehicle data and development with the goal of creating a validated digital twin architecture.

\end{abstract}


\section{Introduction}

Modular electric vehicle platforms with distributed drives and steer-by-wire functionality provide additional degrees of freedom for vehicle dynamics, control development and automated driving research \cite{Hori2004}. Their simulation requires a consistent representation of interacting subsystems such as powertrain, suspension, tires, steering and vehicle body.
The PixKit platform utilized in this work is a modular electric skateboard chassis with all-wheel drive, four wheel-hub motors and selectable steering modes including rear axle steering \cite{PixKit2026}. These features make the vehicle suitable for future investigations of steering strategies, vehicle dynamics and digital-twin-oriented development. The absence of a parameterized full-vehicle model of the platform available in simulation environments like IPG CarMaker was therefore the reason for this project.
This paper presents the development of a first functional and extendable vehicle model of the PixKit platform in IPG CarMaker. The model is not intended as a fully validated high-fidelity representation of the real vehicle. Instead, it is designed as a physically plausible model that integrates all relevant subsystems in a traceable structure. Due to the incomplete data basis, the parameterization combines documented vehicle data, derived values and assumptions.
The resulting model is evaluated through basic simulation scenarios, including straight-line acceleration, minimum turning circle comparison and steady-state circular driving. These tests are used to assess simulation stability, subsystem interaction and qualitative plausibility of the vehicle response. The main contribution of this work is the structured transfer of the PixKit vehicle architecture into an executable CarMaker model that can serve as a low-level basis for future refinement.
The paper addresses the question of how the PixKit platform can be represented in IPG CarMaker as a functional and extendable full-vehicle model based on incomplete parameter data.

\section{Background and Related Work}

\subsection{Digital Twin Concept}

Digital twin technology is increasingly applied in the automotive sector to address rising system complexity while supporting shorter development cycles. Digital twins extend conventional simulation models by linking a virtual representation of a physical system with data from the real object or its operation \cite{Deng2023}. They are not defined by their geometry alone, but by the combination of their physical and virtual model, data exchange and the ability to support analysis and decision making during the vehicle lifecycle \cite{Rjabtsikov2023}.
This distinction is important in vehicle development because a simulation model or virtual prototype can support design decisions before physical testing, whereas a complete validated digital twin requires a structured connection to measurement and sensor data \cite{Rjabtsikov2023}. In the case of electric vehicles, this is especially relevant, since drivetrain, battery, power electronics, vehicle dynamics and control functions are in constant interaction with each other and jointly influence vehicle behavior \cite{Verma2024}.
Therefore, a physics-based vehicle model can be regarded as an enabling layer for a later digital twin architecture, but not as a complete digital twin unless it is calibrated and validated with data from the physical vehicle.

\subsection{Vehicle Modeling and Simulation}

Vehicle models used as a basis for digital twins must reproduce the relevant vehicle behavior with a level of detail that matches the intended development task. In early stages, as in the present work, the objective is often not to reconstruct every physical component in detail, but to define a model structure that includes the essential interactions between body, tires, suspension, steering, braking and powertrain, while remaining transparent and parameterizable for the future \cite{Pinheiro2023}.
Simple bicycle models are useful for describing basic lateral and yaw dynamics with low computational effort, but they cannot represent all subsystems in a full vehicle environment. At the other end of the modeling spectrum, multibody models can describe mechanical structures and kinematic relations in much greater detail, but they require reliable geometry, mass, stiffness and damping data from all over the vehicle. Without such data this type of model adds additional complexity without improving the model’s credibility \cite{Zhang2021,Liu2024}.
Therefore, a parameterized full-vehicle simulation environment is a suitable compromise for the present work. It allows the main vehicle subsystems, driver input, road description and maneuver definition to be represented in one integrated environment, without requiring a full multibody reconstruction. The model can still later be refined and calibrated with validated measurement data of the vehicle \cite{Rjabtsikov2023,IPG2026}.

\subsection{Research Gap and Contribution}
Existing research shows that digital twin concepts are increasingly applied to automotive systems and that electric vehicles are often analyzed through models which break down the vehicle to its subsystems \cite{Deng2023}. However, these studies do not directly cover a transferable model architecture for a modular research platform with incomplete parameter data. This is particularly relevant, since new unconventional vehicle architectures with distributed drive, new steering concepts and modular chassis are playing an important role in modern vehicle development \cite{Deepak2023}. For such platforms, the research need in the field of automated driving systems (ADS) testing is therefore not a highly detailed component model at an early stage, but a comprehensive and extendable vehicle model that can be refined as further measurements become available. The present project addresses this need by establishing a physically plausible full vehicle modeling basis for the PixKit modular electric test platform, which can support early simulation and decision making and serve as a foundation for later digital twin architecture.

\section{Methodology and Vehicle Modeling}

\subsection{Reference Vehicle and Data Basis}
The physical reference system for the CarMaker model is the PixKit modular electric skateboard platform which serves as a research vehicle for the Automotive Safety and Innovation Laboratory Cologne. In the current state of the project, the modeled vehicle represents the configuration of the vehicle with only the chassis and without passenger cabin or additional body structure. This distinction is relevant for the model setup, because the documented curb mass of 420 kg refers to this skateboard configuration, and must be adjusted in later development stages if additional structure changes the mass, center-of-gravity or inertia.
The documentation provided by the manufacturer provides the main geometric reference data used for the initial assembly of the model \cite{PixKit2026}.

\begin{figure}[htbp]
\centering
\includegraphics[width=\columnwidth]{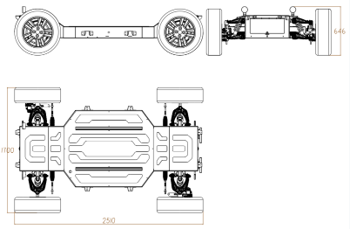}
\caption{PixKit skateboard chassis. Documented dimensions: length 2510 mm, width 1700 mm, height 646 mm, wheelbase 1900 mm, front/rear track width 1465 mm \cite{PixKit2026}}
\label{fig:pixkit_chassis}
\end{figure}

The dimensions seen in Figure \ref{fig:pixkit_chassis} were transferred into the CarMaker coordinate system by calculating the wheel and wheel-carrier positions.
The PixKit architecture combines all-wheel drive, four wheel-hub motors, a simple double wishbone independent suspension, and rear axle steering \cite{PixKit2026}. These features define the main subsystem structure of the CarMaker model consisting of model, chassis, suspension, steering, tires, brakes and powertrain.
All the relevant functions were taken from the manufacturer documentation. However, no extensive measurements have yet been carried out on all components.
The operating range specified in the data set also constrains the intended simulation scenarios. Although the documentation indicates a possible maximum speed of around 60 km/h, the normal operating speed is specified as up to 40 km/h \cite{PixKit2026}. Besides the validation of the maximum possible vehicle straight line speed, maneuvers such as circular driving and cornering are therefore planned within this lower speed range.

\subsection{Modeling Scope and Abstraction Level}
The CarMaker model is developed as a system-level full-vehicle model for low to medium-speed investigations rather than a highly detailed representation of all mechanical, electrical and control subsystems. This abstraction level matches the current research focus of the vehicle platform at the laboratory. PixKit is mostly intended as an experimental platform for future work on autonomous driving functions, sensor integration, steer-by-wire concepts, suspension development and custom passenger cell. High-speed driving and limit-handling behavior, therefore, are not the primary scope of the vehicle.
The objective is to provide a stable and traceable simulation basis which can be extended as soon as more reliable vehicle data becomes available. It is better understood as the first layer of a future digital twin architecture. Its main purpose is to define the subsystem structure and implement all the available data.
To achieve this, several simplifications are introduced. The suspension is currently represented using internal CarMaker suspension data and not by implementing a detailed multibody reconstruction of the real double-wishbone geometry.

Since the suspension is expected to be reworked in future work, detailed measurements of spring and damper characteristics were not the priority. The electric drivetrain is modeled through wheel-based torque generation and drivetrain configuration. In-depth electromagnetic motor behavior, inverter dynamics, thermal effects and advanced battery functions are outside the current modeling scope, although basic high-voltage and low-voltage battery parameters are included to achieve a complete and executable vehicle configuration.
This abstraction level is sufficient for the plausibility assessment of the model and should make the model accelerate consistently, follow basic steering inputs, remain stable and show reasonable longitudinal, lateral and yaw behavior by corresponding to the vehicle’s architecture and geometry.

\subsection{CarMaker Model Architecture}
\begin{figure}[htbp]
\centering
\includegraphics[width=\columnwidth]{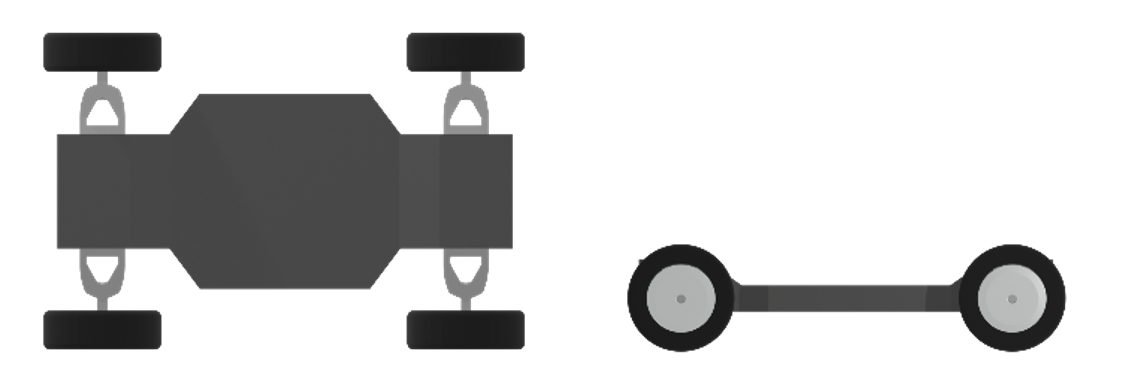}
\caption{Top and side view of the PixKit visualization.}
\label{fig:pixkit_visualization}
\end{figure}

CarMaker provides an integrated simulation environment for virtual test driving, vehicle dynamics, powertrain and system development \cite{IPG2026}.
The vehicle was implemented as a new IPG CarMaker Vehicle Data Set using the CarMaker Office 15.0.1 version. This approach was chosen to avoid hidden assumptions from a predefined vehicle template.
The model follows the standard CarMaker vehicle structure. The simulation relevant parameters are organized in the Assembly, Body, Suspensions, Steering, Tires, Brake and Powertrain sections. Sections that are not required for the current project, for example sensor configuration, vehicle control functions and additional auxiliary settings, were left at their default values or deactivated.
In addition to the simulation-relevant parameterization, a simplified visual model was derived from an available 3D mesh in CATIA and imported into the CarMaker visualization environment after adding textures to the model using Blender (see Figure \ref{fig:pixkit_visualization}).

This visual representation is used for MovieNX and for checking the basic vehicle proportions in the simulation environment. Since CarMaker relies on numerical parameters, it does not define any of the vehicle dynamics and serves only as a geometric and visual reference.

\begin{figure}[htbp]
\centering
\includegraphics[width=\columnwidth]{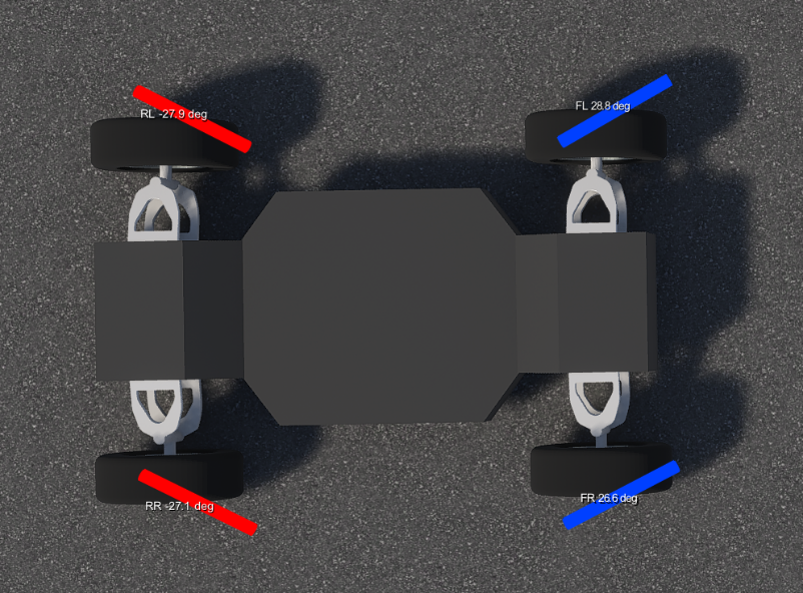}
\caption{Visualization of the PixKit in MovieNX with the additional steering-angle markers.}
\label{fig:pixkit_movienx}
\end{figure}

In order to still allow plausibility checks of the steering implementation, a custom MovieNX Python script was added to visualize the front and rear wheel steering angles during simulation (see Figure \ref{fig:pixkit_movienx}). The script displays colored direction markers at the wheel positions and overlays the corresponding steering angle values.
The project was created with CarMaker for Simulink interface enabled. However, the base vehicle model can be built and simulated independently in CarMaker. The Simulink interface is used as a prepared extension path for the rear axle steering strategy later in the process. This allows the vehicle to be used as a conventional CarMaker model during basic setup work and plausibility checks, while still supporting later integration of the already existing rear axle steering control model.
To keep the paper compact and focus on the simulation results, only the most relevant modeling decisions are described in the following subsections. The complete parameter set, including documented values, assumptions, etc., is stored separately in the project’s Excel file and can be used for later model modifications.

\subsection{Body and Chassis Parameterization}

The body and chassis parameterization defines the geometric and inertial basis of the CarMaker model. The dimensions shown in Figure 1 were used to derive the wheel and wheel-carrier positions for the front axle at x/y/z = 2.205/±0.7325/0.3 m and for the rear axle at x/y/z = 0.305/±0.7325/0.3 m. This preserves the documented wheelbase and track width within the CarMaker reference frame.
The total vehicle mass of 420 kg is divided within the model into a 300 kg rigid body and four wheel-side assemblies of 30 kg each. This distribution is an engineering approximation rather than a measured component breakdown, but it accounts for the wheel-side mass increase caused by the in-wheel motor layout.
The body is modeled as a rigid structure with the center of gravity at x/y/z = 1.255/0/0.323 m. This position is equivalent to the geometric center of the vehicle architecture and results in a 50:50 static axle load distribution. The assumption is acceptable for the current chassis-only model, but must be revised once, for example, the future passenger cell gets installed.
The rotational inertia values are Ixx/Iyy/Izz = 85/170/230 kgm². They were estimated using a simple rigid-body approximation based on the vehicle dimensions and the current mass distribution. These values are preliminary assumptions and are especially relevant for (rear axle) steering investigations, where yaw inertia directly affects the yaw rate response.
A simplified outer shell and basic aerodynamic reference values were added for model completeness, but due to the planned maneuvers at low speed and a small reference area of around 1.1 m², aerodynamic effects are considered less important at this stage.

\subsection{Suspension and Unsprung Mass Modeling}

The real vehicle platform uses independent double-wishbone suspension at the front and rear axle. For the model, this layout is represented by unmodified CarMaker example kinematics using .skc files. These files provide a functional double wishbone representation sufficient for this stage of the digital twin, but do not constitute a measured reconstruction of the actual suspension geometry.
The spring and damper characteristics are modeled with linear elements. Front and rear stiffness are both set to 100 000 N/m using a Hookean spring model. The damping is implemented with a Newtonian damper and identical push and pull damping values of 2500 Ns/m at both axles. These values correspond to the stiff suspension on the real vehicle and at the same time act as placeholders to obtain a stable chassis response during the first simulations. They should not be used for quantitative ride-comfort or handling-limit evaluation.
The increased wheel-side mass caused by the motor configuration is represented through the wheel and wheel-carrier masses. Each wheel is modeled with a mass of 20 kg, including the wheel hub motor, and each wheel carrier with 10 kg, resulting in a total mass of around 30 kg per corner. This is relevant because in-wheel motors shift a larger fraction of the vehicle mass towards the wheel and can therefore influence vertical dynamics significantly.
Some suspension effects are intentionally omitted in the current model. Stabilizers are deactivated by setting their stiffness to 0 N/m, and wheel-bearing flexibility, buffers and parasitic effects are not included. For these reasons the current suspension model remains mostly a placeholder.

\subsection{Powertrain Parameterization}

The powertrain is implemented as an electrical all-wheel drive system with four individual drive sources. Each wheel is assigned to one electric motor. No gearbox is included and the transmission ratio is set to 1, which reflects the physical layout of direct torque application at the wheels without central drivetrain and differential.
For simplification a characteristic value is used to represent the motor characteristics, instead of a measured torque-speed map. The documented rated torque of 126 Nm and rated mechanical power of 3 kW are used for each motor. The maximum rotational speed is limited to 550 rpm, according to the documented peak speed of the powertrain \cite{PixKit2026}. The motor inertia is set to 0.002 kgm², the build-up time to 0.01 s and the friction coefficient to $3 \times 10^{-4}$ Nms/rad as part of a realistic approximation to provide a stable torque response.
Detailed motor current dynamics, inverter switching, thermal behavior and magnetic effects are neglected and will not influence the results of the planned simulations.
Basic supply parameters are included to complete the CarMaker powertrain configuration. According to the documentation, the high-voltage system is set to 72 V and 150 Ah, while the low-voltage system is set to 12 V and 25 Ah \cite{PixKit2026}.
Since the physical model can recuperate energy while braking, a simple generator model is included. The maximum regenerative torque is set to 100 Nm as an assumption, but regenerative braking will not be evaluated in the current model stage.
The chosen architecture keeps the model extendable. Since each wheel already has a separate drive source, measured torque-speed characteristics, efficiency maps, inverter limits and recuperation behavior can be added.

\subsection{Steering and Rear Axle Steering Integration}

The steering system is a central part of the PixKit model because the real vehicle offers three selectable steering modes. Besides conventional front axle steering, the vehicle can be operated in crab steering and counter-phase four-wheel steering. These modes are selected in the physical vehicle by remote control and were therefore implemented as discrete steering configurations rather than as a newly developed rear axle steering controller.
The baseline CarMaker vehicle uses front axle steering only. This allows the basic vehicle model to be tested independently before the rear axle steering function is added. 
The real PixKit platform is documented as a drive-by-wire steering system with Ackermann steering geometry, a maximum inner wheel angle of 30° and a maximum outer wheel angle of 27° \cite{PixKit2026}.
In the current model stage, the actuator dynamics and remote-control hardware are not reproduced in detail.
Rear axle steering is prepared as an extension through the CarMaker for Simulink interface. The existing rear axle steering simulation tool provides a suitable interface because it defines the rear axle steering angle relative to the front axle steering angle \cite{Roehrig2026}. For the present work, this relation is simplified to a mode-dependent constant ratio.

\begin{equation}
k_p=\frac{\delta_r}{\delta_f}
\label{eq:rearsteer}
\end{equation}

For front axle steering, the vehicle model can be used without the extension or with the extension when $k_p = 0$. For crab steering, $k_p=1$ (referring to Parameter A and B = 1 \cite{Roehrig2026}) meaning that front and rear wheels steer in the same direction. For counter-phase steering $k_p=-1$ (referring to Parameter A and B = -1 \cite{Roehrig2026}) meaning that the rear wheels steer opposite to the front wheels.
The expected behavior follows directly from the selected mode. Counter-phase steering reduces the effective turning radius at low speed and is therefore relevant for the minimum turning radius test. Crab steering allows lateral displacement of the vehicle body with reduced yaw motion \cite{ZF2021}.
The front steering mode serves as the reference configuration for comparison. The scope is to evaluate whether the rear axle steering is functionally integrated into the full vehicle model.
The more advanced functions available in the rear axle steering platform, like for example speed-dependent steering characteristics or sideslip compensation, are not used yet since they would not represent the current real vehicle's state.

\subsection{Tire and Brake Modeling}

The tire model was kept deliberately simple, because tire parameter identification is outside the current project scope. Since the exact tire size of the PixKit vehicle 215/45 R17 was not available in the CarMaker presets, all four wheels use the same unmodified RT\_225\_45\_R17. This preset has the same rim diameter and aspect ratio but is 10 mm wider. This increases the nominal sidewall height from 96.75 mm to 101.25 mm and the outer diameter from 625.3 mm to 634.3 mm. The resulting difference in rolling circumference is about 1.4 \%. For this reason, the selected tire preset is only accepted as a practical substitute for basic model checks, since the increased tire width and slightly larger sidewall height may affect lateral force generation and handling behavior.
The PixKit vehicle is documented with a hydraulic drive-by-wire brake system and an electric parking brake \cite{PixKit2026}. In CarMaker this is implemented by a hydraulic basic controller with pressure distribution and pedal actuation. To achieve a typical front-biased braking behavior the torque conversion is set to 6 Nm/bar at the front and to 4 Nm/bar at the rear wheels.
The rear electric parking brake is parameterized by 400 Nm per rear wheel but is not relevant for the planned driving maneuvers.
Since no dedicated braking tests are performed in the current study the brake model is treated as a functional subsystem, rather than replicating the real brake hardware. More detailed brake pressure dynamics, ABS behavior and brake temperature effects remain irrelevant for this project.

\subsection{Plausibility Assessment Strategy}

After the completion of the parameterization, the model is assessed through a set of basic driving scenarios. These scenarios are used to verify whether the implemented subsystems interact consistently and produce plausible vehicle behavior.
A straight-line acceleration test is used to check the longitudinal behavior, consistent function of all four in-hub motors and the achievable maximum speed.
A minimum turning radius maneuver is carried out to compare the front steering baseline with the rear axle steering configuration and is used as a documentation plausibility check against the given steering specifications.
Furthermore, a steady state circular driving test is used to evaluate the influence of the counter-phase rear axle steering on trajectory, yaw rate and lateral acceleration.
The implementation of the rear axle steering is treated as functional integration only. The steering model is parameterized with plausible values based on the available documentation and default values. The main objective is the verification of typical rear axle steering effects, like the reduced turning radius at low speed and the change in yaw response.
If these signals show plausible trends and the simulation remains consistently numerically stable, the model can be considered a functional baseline for its purpose.

\section{Simulation Setup and Plausibility Assessment}

\subsection{Straight-Line Acceleration and Powertrain Response}

The straight-line acceleration scenario was used to assess the basic longitudinal behavior of the implemented vehicle model and the functional integration of the all-wheel-drive powertrain. In this maneuver, the vehicle accelerated from standstill under IPG Driver control, on a straight road, with maximum acceleration input, while rear axle steering was disabled.

\begin{figure}[htbp]
\centering
\includegraphics[width=\columnwidth]{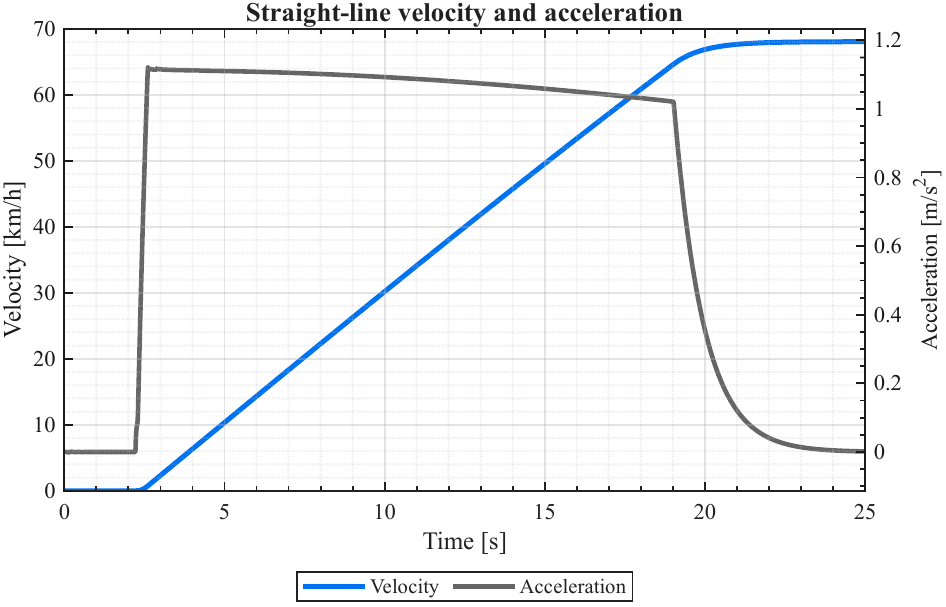}
\caption{Speed and acceleration during the straight-line maneuver.}
\label{fig:speed_acceleration}
\end{figure}

The data from the simulation (see Figure \ref{fig:speed_acceleration}) shows a continuous speed increase up to approximately 68 km/h within about 19 s. The longitudinal acceleration is highest during the initial acceleration phase and decreases as the vehicle approaches its upper speed range, which is consistent with the expected behavior of a power limited electric drivetrain. No unstable oscillations or discontinuities are visible in the speed response.

\begin{figure}[htbp]
\centering
\includegraphics[width=\columnwidth]{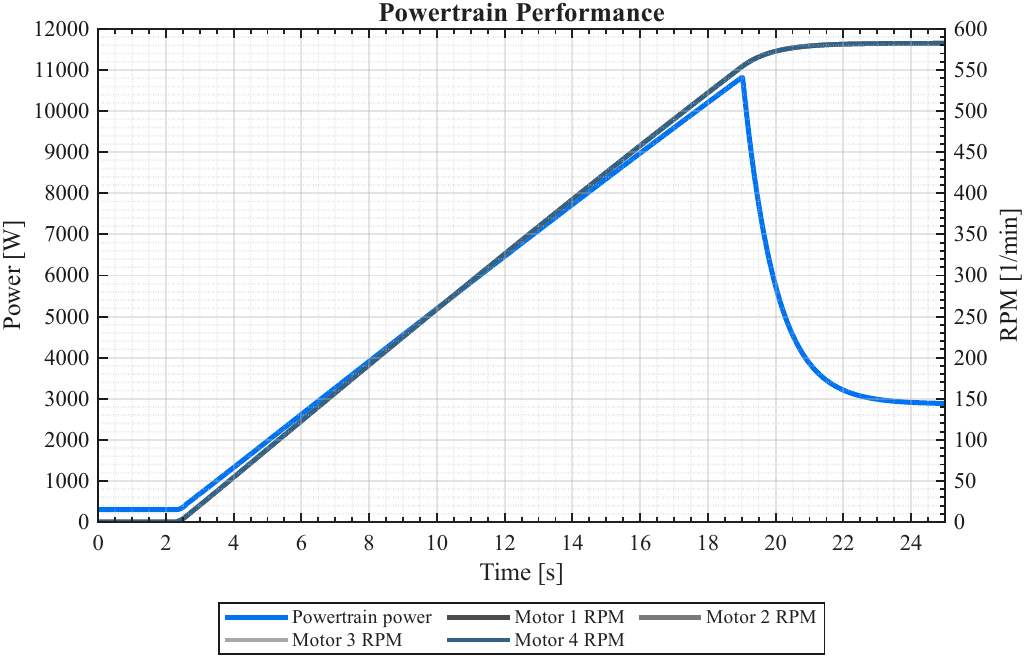}
\caption{Powertrain response during the straight-line maneuver.}
\label{fig:powertrain_response}
\end{figure}

The four motor speed signals remain nearly identical during the whole maneuver (see Figure \ref{fig:powertrain_response}). This behavior is plausible for straight-line driving with symmetric torque distribution and supports the consistency of the implemented powertrain configuration. The total powertrain power peaks around 11 kW and therefore remains close to the documented rated power range of four 3 kW wheel-hub motors but does not exactly reproduce the nominal 12 kW value. Within the limits of the current parameter set, this result is interpreted as a functional plausibility check of the vehicle’s behavior in longitudinal maneuvers.

\subsection{Rear Axle Steering Functionality}
Before evaluating the influence of rear axle steering on the turning trajectory, the implemented rear steering interface was checked using the maximum steering input. The maneuver was performed at 5 km/h with longitudinal motion controlled by the IPGDriver. The documented maximum steering wheel input of 450° was applied with a smooth transition, while the rear axle steering mode was configured in counter-phase steering mode with the given rear to front steering ratio of -1:1.

\begin{figure}[htbp]
\centering
\includegraphics[width=\columnwidth]{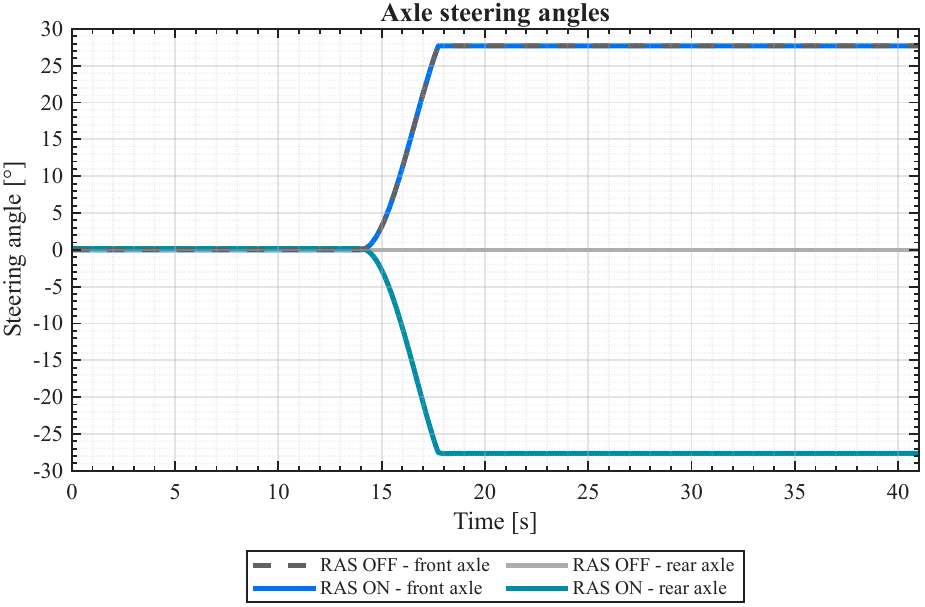}
\caption{Mean axle steering angle during maximum steering input maneuver.}
\label{fig:rear_steering_verification}
\end{figure}

Figure \ref{fig:rear_steering_verification} shows the resulting front and rear axle steering angles for the maneuver with rear steering on (ON) and rear steering off (OFF). For the same maneuver, when rear axle steering is activated, the rear axle follows the front axle with opposite sign and similar magnitude. This confirms that the Simulink-based rear axle steering tool is functionally coupled to the CarMaker model and produces the expected turning angle on the rear axle. In addition, the maximum simulated steering angles were compared with the documented values, which specify a maximum steering angle of 30° for the inner wheel and 27° for the outer wheel. In the model, these values were 28.9° and 26.6°, respectively. The small deviations from the documentation indicate that the implemented steering geometry is close to the real steering range.

\subsection{Minimum Turning Circle With and Without Rear Axle Steering}

The previous test to check the steering functionality was also used to simulate the minimum turning circle with rear axle steering disabled and enabled.

\begin{figure}[htbp]
\centering
\includegraphics[width=\columnwidth]{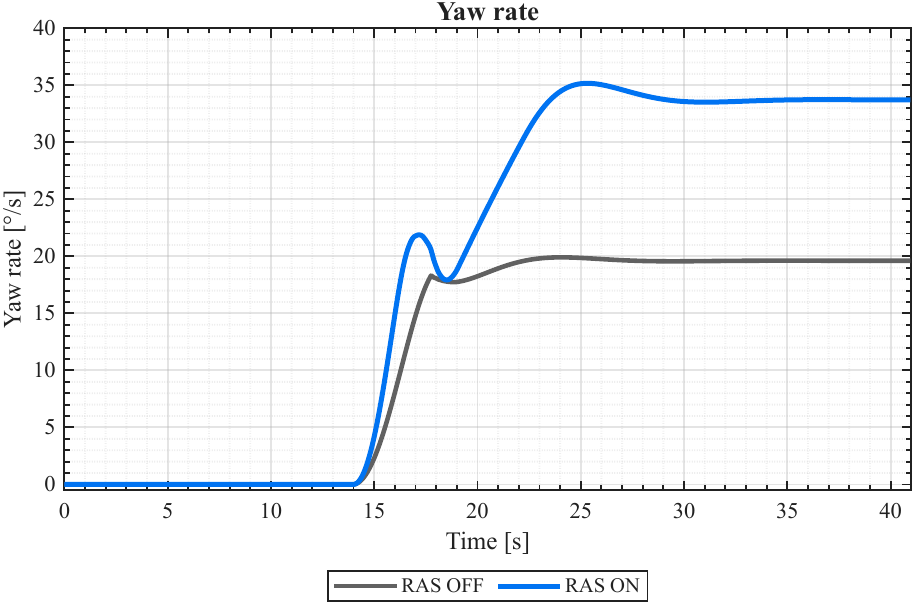}
\caption{Yaw rate comparison during minimum turning circle maneuver.}
\label{fig:yawrate_comparison}
\end{figure}
\begin{figure}[htbp]
\centering
\includegraphics[width=\columnwidth]{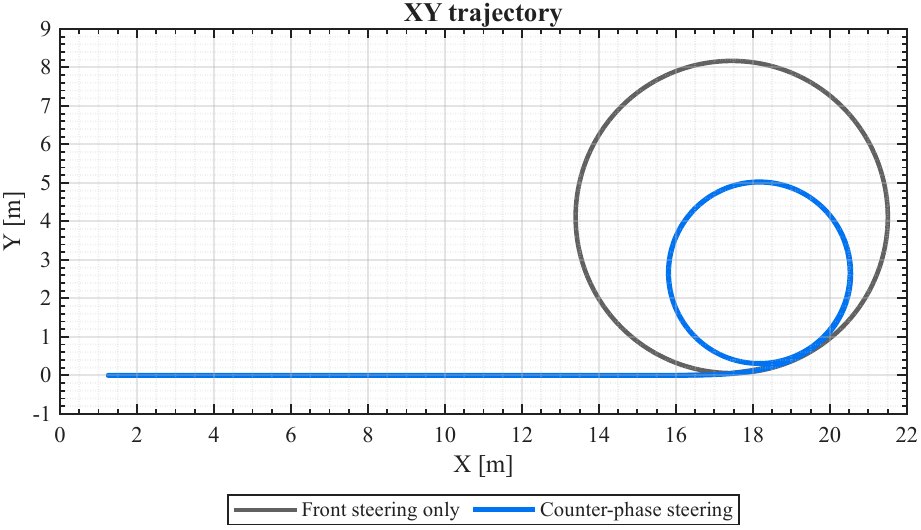}
\caption{Trajectory comparison during minimum turning circle maneuver.}
\label{fig:trajectory_comparison}
\end{figure}

The yaw rate comparison in Figure \ref{fig:yawrate_comparison} shows a higher yaw response when rear axle steering is enabled. This is consistent with the expected effect of counter-phase rear axle steering, where the effective path curvature increases for the same steering-wheel input, shown in Figure \ref{fig:trajectory_comparison}.

The corresponding XY trajectory confirms this behavior on the path level. According to the documentation, the minimum turning circle is 12 m in normal steering mode and 6 m in counter-phase steering mode \cite{PixKit2026}. From the simulated trajectory, the correlating diameters are around 8.2 m and 4.5 m. Although the simulated values are smaller than the documented values, the relative trend is consistent. The active counter-phase steering reduces the simulated turning circle by approximately 45\%, which is close to the documented reduction from 12 m to 6 m.
Since no measurement data from the physical PixKit vehicle were available, the result is not interpreted as a quantitative validation of the real turning radius.

\subsection{Steady-State Circular Driving}

To complete the steering system investigation, the vehicle was put into a steady-state circular driving scenario. The maneuver was performed on a predetermined circular path with the radius set to 20 m and a target speed of 20 km/h. The model completed two full circles each test run, one test run with rear axle steering enabled and one with rear axle steering disabled.

\begin{figure}[htbp]
\centering
\includegraphics[width=\columnwidth]{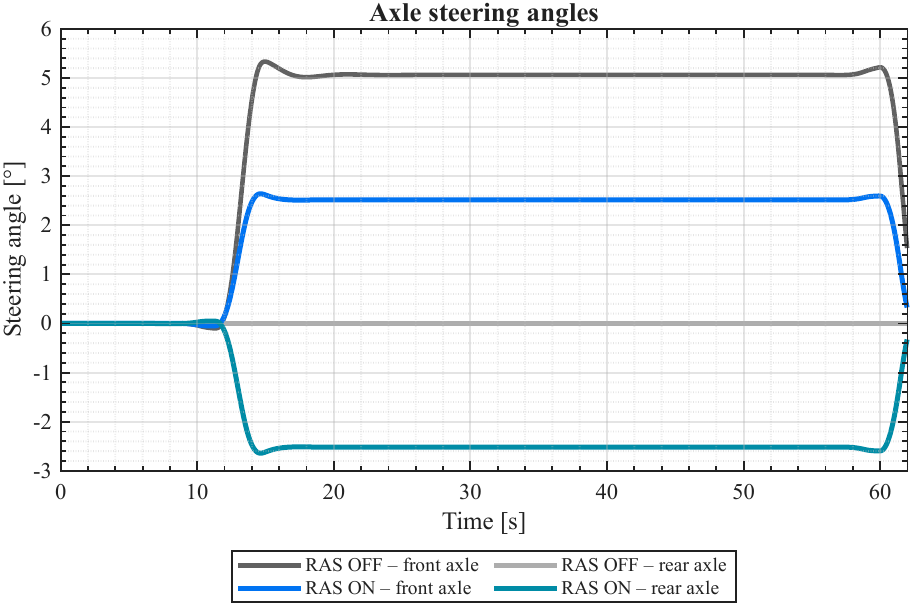}
\caption{Mean axle steering angle during steady-state circular driving.}
\label{fig:circular_steering}
\end{figure}
The axle steering angle comparison in Figure \ref{fig:circular_steering} shows that the baseline model is only governed by the front axle, while the rear axle remains close to zero. With rear axle steering enabled, the steering action is evenly distributed between both axles. The rear axle angle has the opposite sign of the front axle angle, which is consistent with the counter-phase steering mode configuration in the MATLAB/Simulink tool.
\begin{figure}[htbp]
\centering
\includegraphics[width=\columnwidth]{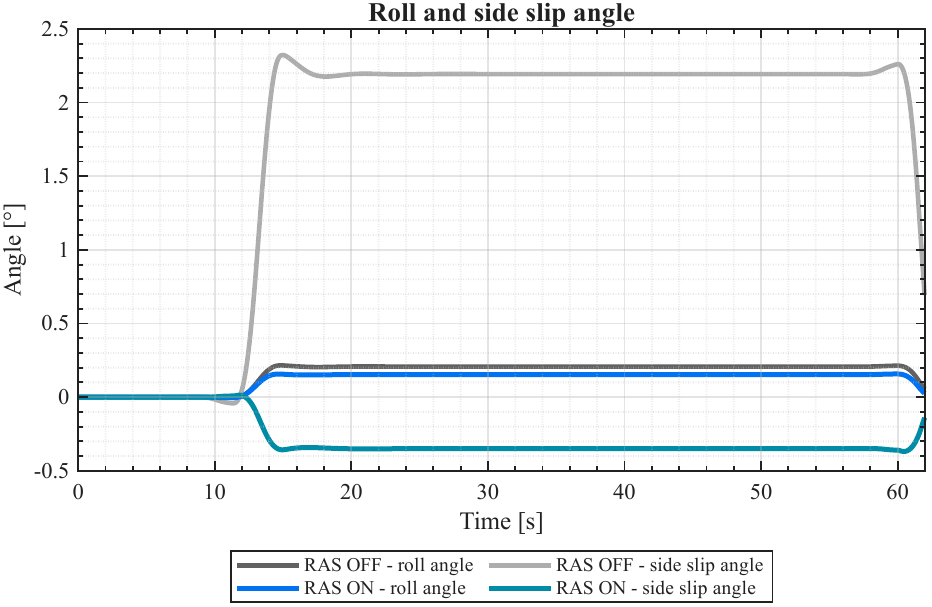}
\caption{Lateral vehicle response during steady-state circular driving.}
\label{fig:lateral_response}
\end{figure}
The roll angle and sideslip angle, as seen in Figure \ref{fig:lateral_response}, remain bounded without visible unstable oscillations. The sideslip angle magnitude is reduced when rear axle steering is on, which is plausible for the selected steering mode, since rear axle steering contributes to the vehicle rotation and changes the orientation of the body relative to the path. The result strengthens the functional assessment of the rear axle steering also during medium-speed maneuvers on a predetermined path.
Additional steering mode tests, such as same-phase steering, were used internally to check flexibility and functionality of the complete rear axle steering interface but are not discussed in detail in this paper.

\section{Discussion}

The simulation results show that the CarMaker model provides a stable and functionally consistent basis for the investigated low- and medium-speed maneuvers. In straight-line acceleration, the vehicle reaches its simulated maximum velocity of 68 km/h within 19 s and shows a symmetric response of the four wheel-hub motors. The total powertrain power reaches 11 kW, which is slightly below the documented rated power of 12 kW for the four motors. This deviation is plausible because the rated power depends on the actual torque-speed operating point, while the simulated acceleration is influenced by the implemented motor speed limitation of 550 rpm as well as internal losses and resistances. However, the simulated final velocity of 68 km/h still exceeds the given documented unloaded maximum speed range of 60/55 km/h \cite{PixKit2026}. Therefore, the model is functionally consistent, but still requires calibration of speed limitation, losses and drivetrain before a quantitative assessment of the real acceleration behavior can be made.

The steering angle results support the plausibility of the implemented steering setup. The PixKit documentation specifies a steering-wheel range of ±450° and maximum inner and outer wheel angles of 30° and 27° \cite{PixKit2026}. While the steering-wheel range of the simulation correlates with the documentation, there are small deviations in the wheel steering angles, with 28.9° and 26.6°.

The minimum turning circle comparison provides the strongest qualitative indication of the rear axle steering effect. With rear axle steering disabled, the maximum yaw rate is approximately 20°/s, while counter-phase rear axle steering increases the maximum yaw rate to around 35°/s under the same steering-wheel input. The produced trajectory shows a reduction of the estimated turning circle diameter from 8.2 m to 4.5 m. This is consistent with the expected low-speed effect of this steering configuration compared to the documented reduction from normal steering to counter-phase steering. However, the simulated diameters were estimated from the vehicle’s center trajectory exported from IPGControl. They are therefore not directly equivalent to a standardized curb-to-curb or wall-to-wall turning circle. This likely explains part of the deviation from the documented turning circle diameters of 12 m and 6 m.

A further limitation concerns the software-side steering behavior of the physical PixKit platform. The documentation mentions a drive-by-wire steering system with target steering-wheel angle and angular velocity as control inputs. In addition, a yaw-rate-related software limit of 40°/s is documented for the vehicle. This limitation was not directly implemented in the current CarMaker steering subsystem, because the present internal model did not provide this implementation. A reproduction of the software-limited yaw behavior would require an additional driver- or controller-level implementation, for example through Simulink. As a result, the yaw-rate results describe the response of the current vehicle model and steering configuration, but not the complete software-limited behavior of the real PixKit vehicle.

The steady-state circular driving scenario further supports the functional plausibility of the model. Since the maneuver was implemented as path following on a predefined 20 m radius at 20 km/h, the resulting steering angles, roll angle and sideslip angle should be interpreted as responses of the model under a defined path-tracking task. The bounded roll and sideslip responses indicate numerical stability and plausible subsystem interaction. Nevertheless, these signals strongly depend on tire, suspension and mass inertia parameters \cite{Pinheiro2023}, which are still based on simplified assumptions and have not been calibrated against measurement data.

\section{Conclusion}

This work presented the conceptual full vehicle modeling of the PixKit modular electric test platform in IPG CarMaker as a first basis for future digital twin architecture. The model represents the current skateboard chassis on system level and includes the main vehicle subsystems required for basic simulation. Since the available vehicle data is incomplete, the parameterization combines documented values, derived parameters, and engineering assumptions.
The conducted simulations show that the model runs stably and produces plausible responses in the investigated simulation environments.
Overall, the current model can be considered a useful low-level simulation basis for future digital-twin-oriented development. It provides a structured integration of the relevant PixKit subsystems in one executable CarMaker model. The results show that the model can reproduce the expected qualitative behavior of the all-wheel-drive powertrain and rear axle steering system.
However, the model remains an initial plausibility-based simulation model and not a quantitatively validated digital twin. Future work should focus on calibrating the model using measurements from real driving maneuvers and connecting the CarMaker model to measured vehicle data and control interfaces.
The research question can therefore be answered in the sense that the PixKit platform can be represented in IPG CarMaker as a functional and extendable full vehicle model by combining manufacturer data, derived parameters and clearly identified engineering assumptions.
\newpage

\bibliographystyle{IEEEtran}
\bibliography{references}

\section*{AI Usage Statement}

During the preparation of this paper, AI-powered language models were used as supporting tools for information overview, language refinement, structural feedback and proofreading. The technical content, modeling decisions, simulations, interpretation of results and final responsibility for the submitted work remain entirely with the author. All AI-assisted suggestions were critically reviewed and adapted by the authors.

\section*{Acknowledgments}

This work was funded by the German Federal Ministry of Research, Technology and Space (Bundesministerium für Forschung, Technologie und Raumfahrt, BMFTR; formerly BMBF) within the funding measure "Forschung an Fachhochschulen – KI-Nachwuchs@FH" under the project "TH Köln – Künstliche Intelligenz plus (THK-KIplus)", funding code 13FH007KI2. The responsibility for the content of this publication lies with the authors.

\newpage


\appendices

\section{Supplementary Project Files}

The submission directory contains supplementary files that provide
additional context and information related to the project.

Link to submission directory: \url{https://git-ce.rwth-aachen.de/asil-lab-cologne/03_publications/02_pixkitcmdigtwin}

\begin{enumerate}

\item The complete parameter documentation is provided as a separate Excel file.
It includes documented manufacturer values, derived parameters, and engineering
assumptions organized according to the internal CarMaker model structure.

\item The complete \textit{Pixkit\_Digital\_Twin\_V1} CarMaker project, including
the following custom-created and modified files:

\begin{enumerate}

\item The parameterized vehicle model
(\textit{PixKit\_V1}),
located in
\textit{Pixkit\_Digital\_Twin\_V1/Data/Vehicle}.

\item Information file used for integrating the 3D vehicle model into CarMaker
(\textit{PixKitV1.assetinfo}),
located in
\textit{PixkitDigitalTwin\_V1/MovieNX/data}.

\item The 3D vehicle model exported from Blender
(\textit{PixKitV1.glb}),
located in
\textit{PixkitDigitalTwin\_V1/MovieNX/data}.

\item Preview images of the vehicle model
(\textit{PixKitV1.preview.png} and \textit{PixKitV1.top.png}),
located in
\textit{Pixkit\_Digital\_Twin\_V1/MovieNX/data}.

\item Python script used for steering-angle visualization in MovieNX
(created analogously to the \textit{.tclgeo} concept used in IPG Movie~\cite{Roehrig2026}),
named
\textit{visualize\_steering\_pixkit.py},
located in
\textit{Pixkit\_Digital\_Twin\_V1/MovieNX/data/Scripts}.

\item The parameterized MATLAB/Simulink model of the rear steering extension~\cite{Roehrig2026},
named
\textit{HAL\_Strategien\_V2\_R2024a.mdl},
located in
\textit{Pixkit\_Digital\_Twin\_V1/src\_cm4sl}.

\end{enumerate}

\item The raw CAD model
(\textit{PixKit\_V1.CATPart}).

\end{enumerate}

\newpage

\section{Model Parameter Overview}

\begin{table}[htbp]
\caption{Overview of Essential Model Parameters}
\label{tab:model_parameters}
\centering
\footnotesize

\begin{tabular}{p{2.8cm} p{3.2cm} p{1.6cm}}
\toprule
Parameter & Value & Source \\
\midrule

Vehicle dimensions & 2510 mm $\times$ 1700 mm $\times$ 646 mm & Documentation \\
Wheelbase & 1900 mm & Documentation \\
Track width & 1465 mm & Documentation \\
Vehicle mass & 420 kg & Documentation \\
Rigid body mass & 300 kg & Assumption \\
Wheel carrier mass & 30 kg per corner (incl. motor) & Assumption \\
Center of gravity & x/y/z = 1.255/0/0.323 m & Assumption \\
Rotational inertia & Ixx/Iyy/Izz = 85/170/230 kgm$^2$ & Assumption \\
Suspension stiffness & 100\,000 N/m & Assumption \\
Damping coefficient & 2500 Ns/m & Assumption \\
Drive concept & Four individual wheel-hub motors & Documentation \\
Rated torque per motor & 126 Nm & Documentation \\
Rated power per motor & 3 kW & Documentation \\
Maximum motor speed & 550 rpm & Documentation \\
HV battery & 72 V / 150 Ah & Documentation \\
LV battery & 12 V / 25 Ah & Documentation \\
Steering modes & Front, crab and counter-phase & Documentation \\
Max. wheel steering angle & 30$^\circ$ inner / 27$^\circ$ outer & Documentation \\
Documented tire size & 215/45 R17 & Documentation \\
CarMaker tire preset & RT\_225\_45\_R17 & CarMaker preset \\
Brake torque conversion & 6 Nm/bar front, 4 Nm/bar rear & Assumption \\

\bottomrule
\end{tabular}
\end{table}

\end{document}